\documentclass[runningheads]{llncs}

\usepackage{eccv}
\usepackage[
  left=1.8in,
  right=1.8in
]{geometry}

\usepackage{eccvabbrv}

\usepackage{graphicx}
\usepackage[table,xcdraw]{xcolor}
\usepackage{colortbl}
\usepackage{tabularx}
\usepackage{booktabs}
\usepackage{caption}
\usepackage{tcolorbox}
\usepackage{siunitx}
\usepackage{amssymb}    
\usepackage{pifont}

\usepackage[accsupp]{axessibility}  

\usepackage{hyperref}

\usepackage{orcidlink}

\begin{document}

\title{Graph it first! Enabling Reasoning on Long-form Egocentric Videos through Scene Graphs} 

\titlerunning{Graph it first!}

\author{
Agnese Taluzzi\inst{1} \and
Riccardo Santambrogio\inst{1} \and
Simone Mentasti\inst{1} \and
Chiara Plizzari\inst{2} \and
Matteo Matteucci\inst{1}
}

\authorrunning{A.~Taluzzi et al.}

\institute{
Politecnico di Milano, Milan, Italy\\
\email{\{agnese.taluzzi,riccardo.santambrogio,simone.mentasti,matteo.matteucci\}@polimi.it}
\and
Bocconi University, Milan, Italy\\
\email{chiara.plizzari@unibocconi.it}
}

\maketitle

\begin{abstract}
Existing multi-modal large language models (MLLMs) face significant challenges in processing long video sequences due to strict input token limitations. As a result, current video understanding approaches, especially in egocentric settings characterized by complex dynamics, frequent state changes, and moving cameras, are forced to massively subsample frames. This leads to severe loss of temporal and contextual information, constraining their ability to perform fine-grained video reasoning. 
In this work, we introduce a framework for egocentric video question answering (VQA) that overcomes these input constraints through \textbf{Egocentric Scene Graphs (EgoSGs)}, i.e., temporally grounded, structured representations that capture objects, attributes, spatial relations, and interactions over time. By representing videos as compact, text-based scene graphs, our method preserves the essential visual and temporal information of the original video in a symbolic form that drastically reduces input length while maintaining semantic richness. Crucially, this enables MLLMs to reason efficiently over entire video sequences within their token budget. 
On \textit{HD-EPIC VQA}, our method achieves state-of-the-art results, outperforming strong video-based baselines on multiple models and suggesting that structured, temporally grounded representations like EgoSGs can bridge long-form egocentric video understanding and the context limitations of today’s MLLMs.
\keywords{Egocentric Vision \and Multi-Modal Large Language Models}
\end{abstract}

\section{Introduction}
\label{sec:intro}
\begin{figure}[t]
    \centering
    \includegraphics[width=\linewidth]{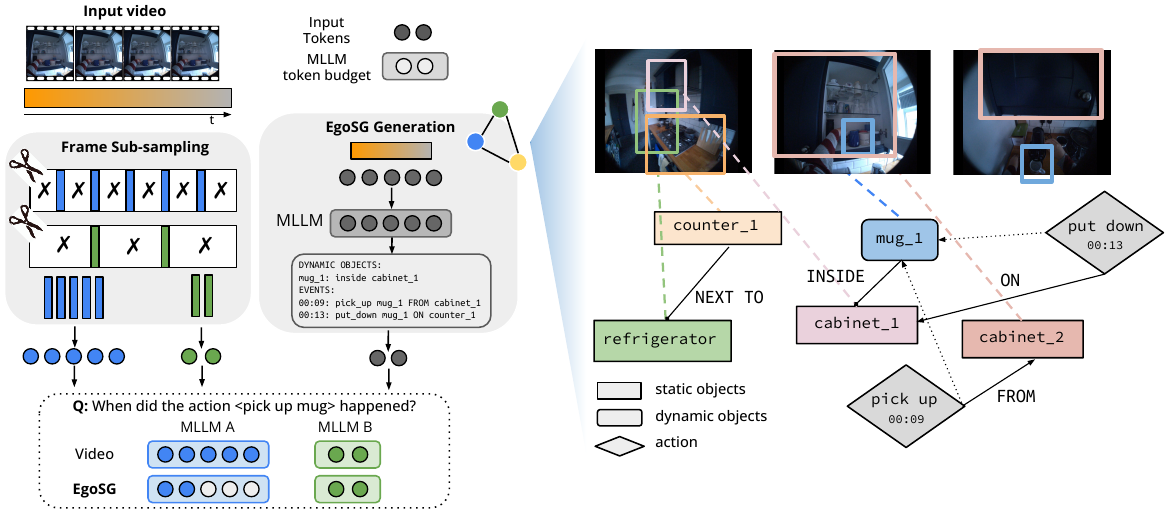}
    \caption{Given a long input video, frames are sub-sampled (blue, green) to respect the \textit{multi-modal large language model (MLLM) token budget}, shown as limited token slots (left). Instead of directly feeding visual tokens, EgoSG generation builds a compact open-vocabulary \textit{scene graph} capturing objects, spatial relations, and actions over time, which is produced in the form of text from an MLLM (right). Nodes represent dynamic (rounded) and static (rectangular) objects, while diamonds denote temporally grounded actions (e.g., \textit{pick up}, \textit{put down}); edges encode spatial (\textit{ON}, \textit{ABOVE}, \textit{NEXT TO}) and interaction (\textit{FROM}, \textit{ON}) relations. The resulting text representation preserves temporal and semantic context while reducing token usage, enabling efficient long-video reasoning on different MLLMs.}
    \label{fig:teaser}
\end{figure}

Egocentric videos are uniquely rich in capturing users’ interactions with objects and their surroundings~\cite{plizzari2024outlook}.
Compared to third-person video understanding, egocentric footage exhibits complex dynamics, frequent viewpoint changes, and long temporal dependencies that demand reasoning across extended time horizons. 
However, current multi-modal large language models (MLLMs), despite their strong perception and reasoning capabilities, are constrained by limited input token capacity. As a result, existing methods must heavily subsample long videos, leading to severe loss of temporal context and incomplete understanding of long-term activities~\cite{yang2025egolife, fu2025video} (see \cref{fig:teaser}).

In this work, we propose a framework that addresses these challenges by incorporating intermediate structured representations that enable fine-grained reasoning over long video sequences without exceeding token limits.
Central to our approach are \textit{scene graphs} specifically designed to capture the dynamic structure of scenes in egocentric video, that we refer to as Egocentric Scene Graphs ({EgoSGs}). Originally developed for static image understanding~\cite{johnson2015image} and later extended to video domains~\cite{xu2017scene, rodin2023actionscenegraphslongform, peddi2025towards}, scene graphs represent temporal dynamics and spatial relations, and interactions in a structured, symbolic format.

\begin{figure}[t]
    \begin{minipage}[t]{0.48\linewidth}
    \vspace{0pt}
        \centering
        \includegraphics[width=\linewidth]{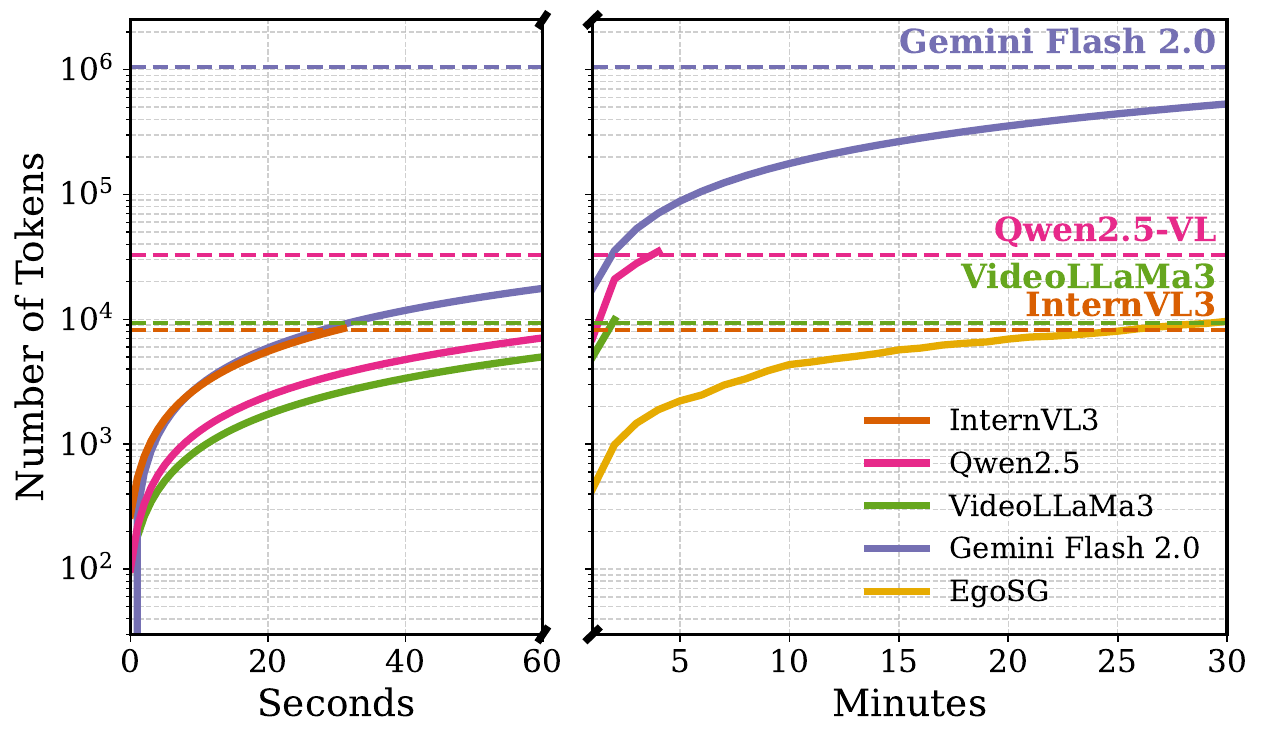}
        \caption{\textbf{Token‐budget comparison.} Cumulative input tokens needed to represent a 1 FPS video across four MLLMs, compared to EgoSG, over short (0–60 s) and long (0–30 min) timescales. Differences in frame-level token counts arise from model-specific tokenizers, whereas EgoSG yields similar token counts across all models. Dashed horizontal lines mark each model’s maximum context window.}
        \label{fig:tokens_per_model}
    \end{minipage}
    \hfill
    \begin{minipage}[t]{0.48\linewidth}
    \vspace{0pt}
        \centering
        \includegraphics[width=\linewidth]{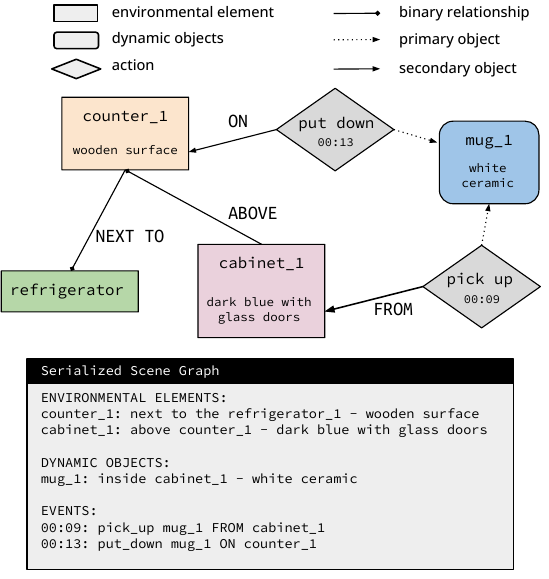}
        \caption{\textbf{EgoSG Schema.} Visualization of EgoSG schema (top) and its serialized textual representation (bottom).}
        \label{fig:egosg-schema}
    \end{minipage}
\end{figure}

We first prompt a large multi-modal language model (MLLM) to extract {EgoSG} representations from egocentric videos in an \textit{open-vocabulary} fashion, capturing the temporal evolution of objects, interactions, and spatial relations in a structured symbolic form. This process yields a compact, text-based intermediate representation that abstracts dense video content into a structured form. Rather than feeding raw video frames directly to MLLMs, we use these scene graphs as the sole input, enabling reasoning over abstracted text instead of low-level pixels. Importantly, while graph generation leverages a large MLLM for high-quality, open-vocabulary parsing, the resulting text-based EgoSGs can subsequently be processed by \textit{smaller} MLLMs for downstream reasoning without frame subsampling. As shown in \cref{fig:tokens_per_model}, traditional frame-level inputs exceed the context budget within seconds or minutes, whereas EgoSG (computed over 1-minute video segments; see \cref{subsec:implementation}) grows slowly and, even after 25 minutes, only reaches InternVL3 and VideoLLaMa3 limits, demonstrating the efficiency of our symbolic intermediate representation.

We validate our approach on the challenging HD-EPIC VQA benchmark~\cite{perrett2025hd}, which contains long-form, egocentric videos requiring multi-step temporal and causal reasoning, showing EgoSG input largely outperforms video-only inputs on multiple MLLMs.  In summary, our key contributions are:
\begin{itemize}
    \item We introduce {EgoSG}, a text-based representation that encodes video content into dynamic scene graphs capturing object interactions, spatial relations, and temporal context.
    \item We propose a pipeline that decouples raw video processing from reasoning by first extracting EgoSGs as an intermediate representation and then feeding only these graphs into off-the-shelf MLLMs, enabling efficient VQA.
    \item We evaluate our framework on the HD-EPIC VQA benchmark, demonstrating state-of-the-art performance over existing MLLMs fed with video input and highlighting the benefits of the proposed graph-based representation.
\end{itemize}

\section{Related Works}
\label{sec:related}
\subsection{Egocentric VQA}
While egocentric video has gained significant attention in recent years, few datasets are explicitly designed to address the challenges of egocentric VQA, and existing benchmarks fall short in evaluating long-form temporal reasoning. Early efforts such as EgoVQA~\cite{fan2019egovqa} and EgoTaskQA~\cite{jia2022egotaskqa} focus on coarse-level questions or short-term action dependencies, often relying on synthetic or weakly grounded annotations that allow models to exploit commonsense priors rather than genuine temporal reasoning~\cite{plizzari2025omnia}.
SGQA~\cite{cho2026perceptionlm} captures real-world smart glasses data but remains small in scale ($<$6 hours). More recent efforts build on large-scale datasets like Ego4D~\cite{grauman2022ego4d}, including EgoSchema~\cite{mangalam2023egoschema} and QAEgo4D~\cite{barmann2022did, patel2025advancing}. 
However, existing benchmarks consist of relatively short clips, for example, EgoSchema (avg $\sim$3 minutes) and EgoTaskQA (avg $\sim$25 seconds), which limits their ability to evaluate how severe frame subsampling affects overall model performance.

We evaluate our approach on the \textit{HD-EPIC} benchmark~\cite{perrett2025hd}, a large-scale dataset specifically designed for fine-grained reasoning in egocentric settings. Unlike prior benchmarks composed of short, trimmed clips, \textit{HD-EPIC} features long, untrimmed videos (averaging 44.6 minutes per question), making it uniquely suited to assess models’ ability to reason over extended temporal contexts. While long-form video QA also exists in exocentric settings, our focus is egocentric; extending the approach to datasets with multiple interacting agents and exocentric dynamics is non-trivial, and we leave it to future work.

\subsection{Multi-Modal LLMs}
\begin{sloppypar}
Large Language Models (LLMs) such as GPT-4~\cite{openai2023gpt4} and LLaMA \cite{llama2023llama} have demonstrated impressive generalization and reasoning capabilities across a wide range of text-based tasks. Building on this foundation, recent models like GPT-4V(ision)~\cite{openai2023gpt4v} and Gemini~\cite{team2024gemini} extend these capabilities to the visual domain, enabling multi-modal reasoning over both text and images. This has led to the emergence of Multi-Modal Large Language Models (MLLMs) such as BLIP~\cite{li2022blip}, Flamingo~\cite{alayrac2022flamingo}, Kosmos~\cite{peng2024grounding}, PaLM-E~\cite{driess2023palm}, and others~\cite{huang2023language}, which achieve strong results across vision-language benchmarks. 
More recently, MLLMs have been adapted to video understanding, with models such as Gemini~\cite{team2024gemini}, GPT-4o~\cite{openai2024gpt4o}, Qwen-VL~\cite{wang2024qwen2, bai2025qwen25vltechnicalreport}, VideoLLaMA~\cite{zhang2025videollama}, Video\-LLaVA~\cite{maaz2024video}, MiniCPM\-V~\cite{yao2024minicpm}, InternLM\-XComposer~\cite{dong2024internlm}, and LLaVA\-Next~\cite{liu2024llavanext} incorporating multi-frame context and temporal reasoning modules.
\end{sloppypar}
However, despite improvements in context length, e.g., GPT-4o supports $\sim$128K tokens and Gemini up to two million tokens, the number of frames that can be effectively processed remains limited. When video frames are tokenized into visual embeddings, even high-capacity MLLMs can handle only a few hundred frames, corresponding to less than a few minutes of footage. As a result, models must heavily subsample long videos, discarding the majority of temporal information and losing continuity in actions and object interactions. Moreover, reasoning directly on pixel-level inputs introduces redundancy and hinders the modeling of higher-level temporal and semantic structure~\cite{plizzari2025omnia}. 

To overcome these limitations, we propose to extract intermediate Egocentric Scene Graph (EgoSG) representations that capture objects, attributes, spatial relations, and interactions over time. By feeding these compact, structured representations to MLLMs, our framework directly addresses the token bottleneck and enables faithful long-term reasoning across entire video sequences in a token-efficient and interpretable manner, eliminating the need for aggressive frame subsampling and outperforming direct video-to-MLLM approaches.

\subsection{Scene Graphs for Video Understanding Tasks}  
Scene graphs have emerged as effective intermediate representations to bridge raw visual input and structured semantic understanding in video analysis~\cite{arnab2021unified,ji2020action,mao2022dynamic,rai2021home,zhao2023constructing}. For instance,~\cite{arnab2021unified} proposed a unified transformer model that jointly reasons over video frames and scene graphs for improved spatio-temporal grounding. The Action Genome dataset~\cite{ji2020action} provides dense scene graph annotations to model human-object interactions and improve action detection. Other works have explored dynamic scene graphs to capture temporal changes in object relations~\cite{mao2022dynamic,zhao2023constructing}. In first-person settings, the egocentric viewpoint introduces unique challenges and opportunities for capturing interactions grounded in the camera wearer’s perspective. Prior work has demonstrated the benefits of structured graph-based representations in this domain.
In~\cite{rai2021home}, hierarchical graph structures tailored to egocentric videos are employed for predicting future actions.
Ego-Topo~\cite{nagarajan2020ego} focuses on spatial topology in egocentric videos to support long-term understanding and action anticipation, though it does not explicitly encode object-relationships. EASG~\cite{rodin2024action} defines a time-varying graph with nodes for the camera wearer, objects, and actions, and edges capturing fine-grained temporal and relational dependencies such as with, from, or direct object. In a concurrent work, \cite{chu2025fine} explores the use of scene graph for improving on a captioning task. 

In this work, we use scene graphs as intermediate representations for egocentric VQA.
However, our approach differs fundamentally from prior methods that rely on manually annotated or task-specific scene graphs~\cite{ji2020action, rodin2024action}.
By leveraging an MLLM, EgoSG automatically generates scene graphs in zero-shot, open-vocabulary manner, without requiring any ground-truth annotations. 
We show that these general-purpose structured abstractions significantly enhance the ability of MLLMs to efficiently perform reasoning over long time horizons.

\section{Our Approach: EgoSG}
\label{sec:method}
We replace video inputs with {MLLM-extracted, text-based Egocentric Scene Graphs (EgoSGs)}. Given a video, a multi-modal LLM is prompted to parse each video into a temporally grounded graph (serialized as text). The resulting textual EgoSG preserves semantics across all frames while removing redundancy, allowing downstream MLLMs to reason over the full sequence within input-token limits. We define the task in \cref{subsec:task}, describe the graph generation process in \cref{subsec:generation}, present the prompting schema in \cref{subsec:schema}, and outline post-processing steps and the MLLM inference in \cref{subsec:post} and \cref{subsec:inference}.

\begin{table}[t]
\centering
\caption{\textbf{EgoSG schema.} Structured overview of EgoSG components, types, and representative attributes.}
\scriptsize
\begin{tabularx}{\linewidth}{lX}
\toprule
\textbf{Component Category} & \textbf{Types and Attributes} \\
\midrule

\multicolumn{2}{l}{\textbf{Nodes ($N$)}} \\
\textbf{Node Types} &
\texttt{Dynamic Object}, \texttt{Environmental Element} \\
\textbf{Node Attributes} &
\texttt{element\_id} (label), \texttt{V} (visual properties) \\
\midrule

\multicolumn{2}{l}{\textbf{Binary Relationships ($E_B$)}} \\
\textbf{Relation Types} &
\texttt{INSIDE}, \texttt{ON}, \texttt{UNDER}, \texttt{NEXT\_TO}, etc. (expressed in free-form natural language) \\
\midrule

\multicolumn{2}{l}{\textbf{Action Hyperedges ($A$)}} \\
\textbf{Action Types} &
Free-form verbs describing user-object interactions, e.g., \texttt{open}, \texttt{pour\_FROM}, \texttt{place}, \texttt{take} \\
\textbf{Main Object} &
Primary entity involved in the action (from \texttt{Dynamic Object} or \texttt{Environmental Element}) \\
\textbf{Secondary Objects} &
Entities affected or referenced by the action (from \texttt{Dynamic Object} or \texttt{Environmental Element}) \\
\textbf{Temporal Scope} &
Single timestamp or interval \( (t_s, t_e) \) denoting when the action occurs \\
\bottomrule
\end{tabularx}
\label{tab:egosg-schema}
\end{table}

\subsection{Task Definition}
\label{subsec:task}
Formally, given a video \( V \) and a natural language question \( Q \) from the set of question types, the goal is to predict the correct answer \( a^* \) from a candidate answer set \(\mathcal{A} = \{a_1, a_2, a_3, a_4, a_5\} \). The model must learn a mapping $f:(V, Q) \rightarrow a^*$, where \(f\) effectively integrates temporal and spatial cues from the video with semantic understanding of the question to select the correct answer.

\subsection{EgoSG Generation}
\label{subsec:generation}
MLLMs have become a popular choice for scene graph generation in recent times, given the large amount of pre-trained knowledge they have~\cite{chen2023gpt4sgg,xu2025llava,li2024pixels}. 
Given an input video $V$ of total duration $T$, we partition it into a sequence of non-overlapping temporal chunks of fixed length $\Delta t$ (e.g., $\Delta t = 60$ seconds), resulting in $N = \left\lfloor \frac{T}{\Delta t} \right\rfloor$ non-overlapping segments: 
\begin{equation}
    V = \{v_1, v_2, \dots, v_N\}, \quad \text{with } v_i \in [t_i, t_i + \Delta t)
\end{equation}
For each segment $v_i$, we generate a corresponding scene graph $\mathcal{G}_i$ using a prompted MLLM, following the schema described in \cref{subsec:schema}. The scene graphs are constructed in a temporally iterative fashion: the initial graph $\mathcal{G}_1$ is generated from $v_1$, and for each subsequent segment $v_{i}$ ($i > 1$), the scene graph $\mathcal{G}_{i}$ is obtained by prompting the MLLM to update $G_{i-1}$ based on new observations in $v_i$. That is,
\begin{equation}
    \mathcal{G}_i = \text{Update}(\mathcal{G}_{i-1}, v_i)
\end{equation}
where $\text{Update}(\cdot)$ denotes a prompted reasoning step guided by the MLLM that incrementally modifies the previous graph to reflect the new visual context. In particular, the initial prompt instructs the model to extract environment elements, dynamic objects, and actions from the first clip, while subsequent prompts explicitly condition on the accumulated context from prior clips. This design ensures continuity across time, prevents redundant listings of previously observed entities, and encourages the model to focus on novel objects and interactions in each new segment. The supplementary material provides the exact prompts employed for both the initial graph generation and the subsequent update steps.

This process produces a temporally-evolving sequence of scene graphs $\{\mathcal{G}_1,\allowbreak \mathcal{G}_2,\allowbreak \dots,\allowbreak \mathcal{G}_N\}$ that captures the dynamic structure of the egocentric scene over time.  At the final timestep, the graph $\mathcal{G}_N$ serves as a compact representation of the entire video $V$.

\subsection{EgoSG Schema}
\label{subsec:schema}
An Egocentric Scene Graph $\mathcal{G}$ is structured as $\mathcal{G} = (N, E_B, A)$, where $N$ is the set of nodes (entities), $E_B$ is the set of binary edges representing direct structural relationships, and $A$ is the set of hyperedges representing interactions of the user with objects in the video. The conceptual design of the entity set $N$ and its associated structural relations $E_B$ draws inspiration from 3D scene representations~\cite{armeni20193d}, here adapted to 2D video. The use of hyperedges in $A$ to model interactions follows from recent work on situation hypergraphs for multi-entity video understanding~\cite{urooj2023learning}.

The structure and content of $N$, $E_B$, and $A$ are described below and illustrated in \cref{fig:egosg-schema}, with their examples summarized in \cref{tab:egosg-schema}.

\subsubsection{Core Entities (Nodes, \( N \)).}  
The node set \( N = \{n_1, n_2, \dots, n_{|N|}\} \) includes all identifiable entities within the scene. These comprise both \textit{dynamic objects} (entities with which the user can interact) and \textit{environmental elements}, which refer to static components of the environment. Each node \( n \in N \) is associated with a label (\texttt{element ID}) and an attributes \( V \) that describe its visual properties.

\subsubsection{Direct Relationships (Binary Edges, \( E_B \)).}  
The set \( E_B \subseteq N \times R_B \times N \) defines the directed binary relationships between pairs of nodes, where \( R_B \) is the set of binary relation types. Each edge \( e_b = (n_i, r_b, n_j) \) expresses that node \( n_i \) stands in a \textit{location-based relationship} \( r_b \) with node \( n_j \). These relationships are described in free-form natural language and capture either the spatial configuration of environmental elements relative to each other or the initial position of dynamic objects with respect to the surrounding environment, e.g. \texttt{milk\_bottle\_1 INSIDE fridge\_1}. 

\subsubsection{Interactions (Action Relationships / Hyperedges, \( A \)).}  
We define the interaction set as \( A = \{a_1, a_2, \dots, a_{|A|}\} \), where each element \( a \in A \) is a hyperedge representing a temporally localized, multi-entity interaction. Each interaction is defined as $a = (\texttt{action}, \texttt{preposition(s)}, \mathcal{M}_a, \mathcal{S}_a, t)$, where:

\begin{itemize}
    \item \texttt{action} is a semantic verb label (e.g., \texttt{pour}) denoting the type of interaction occurring between entities.

    \item \texttt{prepositions} is a set of typed relational markers (e.g., \texttt{FROM}, \texttt{INTO}) that define directed roles between the primary object and one or more secondary objects.

    \item \( \mathcal{M}_a = n_k \) denotes the \textit{primary object} associated with action \( a \) (e.g.,  \texttt{pour}  \texttt{\underline{milk}} \texttt{FROM}, \texttt{INTO})

    \item \( \mathcal{S}_a = \{ (p_i, n_j) \} \) is a set of \textit{secondary object} tuples, where each \( (p_i, n_j) \) binds a secondary object \( n_j \) to the primary object via preposition \( p_i \), specifying its role (e.g.,  \texttt{pour}  \texttt{{milk}} \texttt{FROM} \texttt{\underline{bottle}} \texttt{INTO} \texttt{\underline{glass}}).

    \item \( t \) denotes the temporal extent of the interaction, expressed either as a single timestamp or an interval \( (t_s, t_e) \).
\end{itemize}

\subsection{Post-processing}
\label{subsec:post}
We serialize each scene graph as a structured textual representation. The output begins with a listing of all static environment elements and dynamic objects observed in the scene, using the format: \texttt{ENVIRONMENT ELEMENTS: \{...\}}, \texttt{DYNAMIC OBJECTS: \{...\}}. Each entry includes object identifiers, their location-based relations, and a brief visual description. Subsequently, we describe interactions in the form of timestamped events under the section \texttt{EVENTS}. Each event is written as a line in the format: \texttt{timestamp: action relationship}, where relationships denote the interaction between agents and objects (e.g., \texttt{pick\_up mug\_1 FROM cabinet\_1}). An example of serialized graph is shown in \cref{fig:egosg-schema}.

To reduce redundancy and improve clarity, consecutive identical actions are merged into a single entry using a timestamp range notation \((t_s, t_e)\), preserving both temporal dynamics and readability.

\subsection{MLLM Inference}
\label{subsec:inference}
During inference, the MLLM receives as input the natural language question \( Q \), along with the encoded graph representation \( \mathcal{G} \) generated by the EgoSG pipeline, which substitutes the raw video \( V \). The graph \( \mathcal{G} \) serves as a compact and structured summary of the dynamic video content, capturing key spatial and temporal relationships among entities.
Based on this input, the MLLM selects the most likely answer \( a^* \) from the candidate set \(\mathcal{A}\) via the mapping $f:(\mathcal{G},Q)\rightarrow a^*$.
This approach enables the model to integrate the question's semantic understanding with structured scene information.

\section{Experiments}
\label{sec:experiments}
\subsection{Dataset}
\label{subsec:dataset}
Our evaluation leverages the rich annotations of the HD-EPIC dataset~\cite{perrett2025hd}, focusing on five key question types: \textit{Recipe} localization and recognition, \textit{Ingredient} identification including amounts and order, fine-grained \textit{Action} details, \textit{3D Perception} of object positions, and \textit{Object Motion} tracking over time. Categories concerning nutritional content and gaze estimation are excluded for our evaluation as they do not inherently require long-term temporal reasoning. For instance, the \textit{Nutrition} category includes questions such as ``What is the ingredient with the highest carbs in this recipe?'', which can be answered through commonsense or static attribute reasoning rather than temporal understanding. Similarly, the \textit{Gaze} category focuses on questions like ``What is the person looking at in this segment?'', which depend on fine-grained visual cues and gaze direction, features that are not explicitly captured in the graph-based abstraction. We further validate this choice empirically in \cref{subsec:results}.
Each question is formulated as a 5-way multiple choice using 30 different question templates (or prototypes). Some questions require reasoning about specific objects and time intervals in the video. For example, a question like ``How many times did the object \texttt{<bbox>} seen at \texttt{<time>} move in the video?'' demands tracking the referenced object across frames and counting its movements, demonstrating the need for fine-grained spatio-temporal reasoning.
For our experiments, we select 50 Q\&As per question prototype, for a total of 1250 questions.
This ensures a balanced evaluation while reducing computational overhead.

\subsection{Implementation details}
\label{subsec:implementation}
We use a powerful model, Gemini Flash 2.0, to generate the EgoSGs, which then enables smaller MLLMs to reason over these compact, text-based graphs. We also attempted to use open-source models for scene-graph generation; nonetheless, even the latest open-source MLLMs struggle to reliably produce accurate graphs.
An analysis of the quality of the generated graphs can be found in Supp. Videos are divided into non-overlapping 60-second clips and processed at 1 FPS. We selected 60-second segments based on empirical evaluation, as this duration balances temporal detail and processing efficiency (see ablation in the Supp.). For each 1-minute segment, we generate both a serialized scene graph and a textual summary; details of the prompts are provided in the Supp.
We compare two input configurations for VQA inference: raw video and EgoSG. In both configurations, if the question explicitly contains visual references (e.g., bounding boxes or frame snippets), we extract the referenced images from the video and provide them alongside the input. For the video configuration, raw frames are sampled at 1 FPS. For EgoSG, the video input is replaced by the serialized scene graph.
When the concatenated scene graphs fit within the model’s context window, we use them as input; if they exceed the context limit, we instead feed the concatenated per-segment summaries (see details in Supp.). Our ablation (\cref{tab:results_by_format_and_category}) shows this causes only a minor drop in overall performance.

\begin{table}[t]
\centering
\caption{\textbf{EgoSG Results.} Accuracy (\%) comparison between EgoSG and raw video baseline. Best result per category is \textbf{bolded}. Performance deltas in the \textit{Overall} column show improvement (\textcolor{green!60!black}{$\uparrow$}) or degradation (\textcolor{red}{$\downarrow$}) when using EgoSG.}
\label{tab:egosg}
\resizebox{\textwidth}{!}{
\setlength{\tabcolsep}{4pt} 
\begin{tabular}{ll l ccccc >{\columncolor[gray]{0.95}}c}
\toprule
\textbf{Model} & \textbf{Dim.} & \textbf{Input} & \textbf{Recipe} & \textbf{Ingred.} & \textbf{Action} & \textbf{3D Perc.} & \textbf{Object Mot.} & \textbf{Overall} \\
\midrule
Gemini Flash 2.0 & -- & Videos & 51.75 & 34.00 & \textbf{46.00} & 39.00 & 26.67 & 39.48 \\
 & & EgoSG & \textbf{61.50} & \textbf{37.33} & 38.00 & \textbf{44.00} & \textbf{32.67} & \textbf{42.70} \textcolor{green!60!black}{\small$\uparrow$3.22} \\
\midrule
Qwen2.5-VL & 3B & Videos & 37.75 & 29.67 & 31.00 & 30.50 & 26.67 & 31.12 \\
 & & EgoSG & \textbf{42.25} & \textbf{31.67} & \textbf{33.00} & \textbf{39.00} & \textbf{30.00} & \textbf{35.18} \textcolor{green!60!black}{\small$\uparrow$4.06} \\
\cmidrule(lr){2-9}
 & 7B & Videos & 37.25 & \textbf{30.67} & 32.00 & 25.50 & \textbf{38.00} & 32.68 \\
 & & EgoSG & \textbf{46.25} & 27.00 & \textbf{34.50} & \textbf{37.00} & 34.67 & \textbf{35.88} \textcolor{green!60!black}{\small$\uparrow$3.20} \\
\midrule
InternVL3 
 & 2B & Videos & \textbf{35.00} & \textbf{30.67} & \textbf{33.50} & 24.00 & 26.00 & \textbf{29.83} \\
 & & EgoSG & 31.00 & 20.00 & \textbf{33.50} & \textbf{34.00} & \textbf{28.67} & 29.43 \textcolor{red}{\small$\downarrow$0.40} \\
\cmidrule(lr){2-9}
 & 8B & Videos & 35.00 & \textbf{29.00} & \textbf{40.00} & 31.00 & 20.00 & 31.00 \\
 & & EgoSG & \textbf{42.25} & 18.67 & 33.00 & \textbf{38.50} & \textbf{34.00} & \textbf{33.28} \textcolor{green!60!black}{\small$\uparrow$2.28} \\
\cmidrule(lr){2-9}
 & 14B & Videos & 37.00 & \textbf{36.00} & 38.00 & 35.50 & 18.67 & 33.03 \\
 & & EgoSG & \textbf{53.50} & 28.00 & \textbf{40.50} & \textbf{40.50} & \textbf{30.67} & \textbf{38.63} \textcolor{green!60!black}{\small$\uparrow$5.60} \\
\midrule
VideoLLaMa3 & 2B & Videos & \textbf{31.75} & \textbf{30.67} & 25.00 & \textbf{29.00} & \textbf{30.00} & \textbf{29.28} \\
 & & EgoSG & 28.25 & 20.33 & \textbf{27.00} & 28.00 & 26.00 & 25.92 \textcolor{red}{\small$\downarrow$3.36} \\
\cmidrule(lr){2-9}
 & 7B & Videos & \textbf{45.50} & 31.33 & 32.50 & 31.00 & 28.67 & 33.80 \\
 & & EgoSG & 39.75 & \textbf{32.00} & \textbf{33.00} & \textbf{35.50} & \textbf{30.67} & \textbf{34.18} \textcolor{green!60!black}{\small$\uparrow$0.38} \\
\bottomrule
\end{tabular}
}
\end{table}

\label{subsec:results}
\begin{figure}[t]
    \centering
    \includegraphics[width=\linewidth]{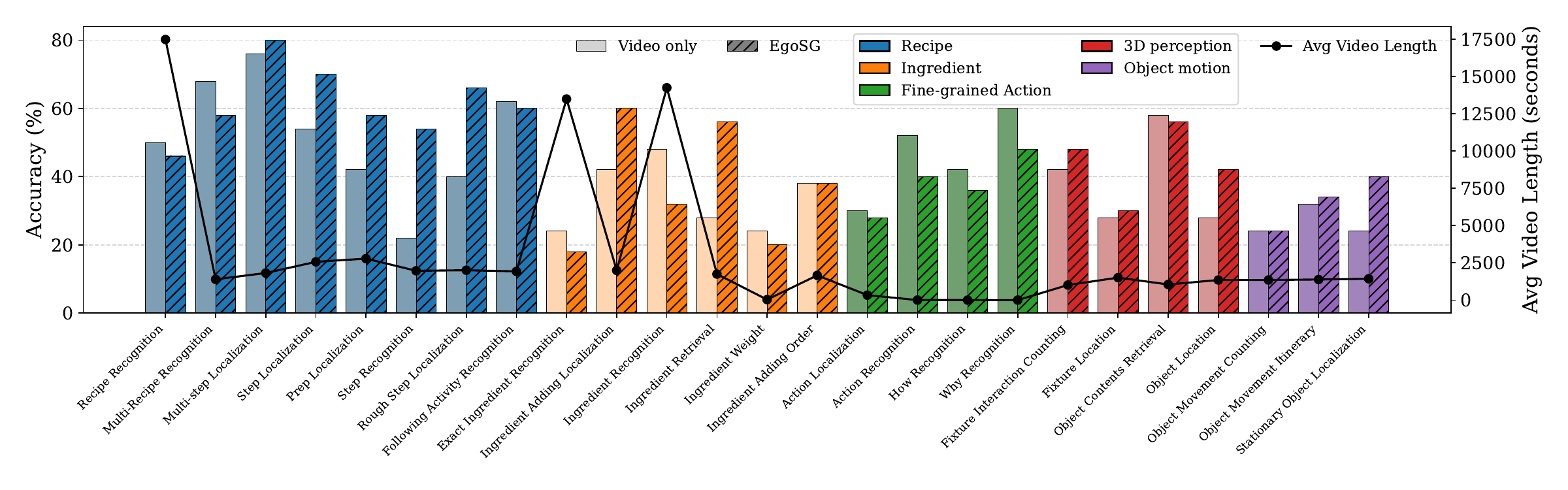}
    \caption{\textbf{Per-prototype Results.} Accuracy (\%) of EgoSG vs raw video input for each question prototype.}
    \label{fig:prototype}
\end{figure}

\subsection{EgoSG Results}
We evaluate our approach using Gemini Flash 2.0~\cite{GeminiFlash}, InternVL3 (2B/8B/14B)~\cite{zhu2025internvl3}, Qwen2.5-VL (3B/7B)~\cite{bai2025qwen25vltechnicalreport}, and VideoLLaMA3 (2B/7B) \cite{zhang2025videollama}. \cref{tab:egosg} compares the performance of models using raw video input versus EgoSG. Our raw video baseline is consistent with the evaluation method used in the original HD-EPIC paper~\cite{perrett2025hd}, which also evaluated models directly on video inputs.

Averaging across all categories (\textit{Overall}), EgoSG consistently outperforms video input for most models, except for smaller models (InternVL3-2B and VideoLLaMa3-2B). The largest improvement is observed for Qwen2.5-VL 14B, with a gain of 5.6\%. This suggests that as model capacity increases, the compact semantic information of a scene graph becomes a more efficient input than the high redundancy of raw video frames.
Critically, EgoSG improves performance even for Gemini (+3.22\%), the same model used to generate the scene graphs, confirming that the gains are driven by the intermediate EgoSG representation itself rather than by a stronger MLLM backbone.

The largest category-level gains appear in \textit{Recipe} (+16.5\%), \textit{3D Perception} (+11.5\%), \textit{Object Motion} (+14.0\%), and \textit{Ingredient} (+3.3\%), which are the tasks that often involve long videos with complex spatial reasoning or multiple object interactions.  Conversely, EgoSG performs worse in \textit{Action}, which involves short video clips lasting a few seconds, and where short-term visual dynamics can favor the raw video signal. Overall, EgoSG offers a significant advantage for tasks demanding structured reasoning over extended interactions.

\cref{fig:prototype} shows performance of EgoSG \textit{vs} raw video across different question prototypes. Within categories, results vary across question prototypes, reflecting the diversity of visual and reasoning demands. Prototypes requiring precise visual details (e.g., \textit{Action Recognition}) tend to favour video-only input, while those involving sequential reasoning or object tracking over time (e.g., \textit{Recipe Step Recognition}) benefit most from EgoSG.

\begin{figure}[t]
    \begin{minipage}[t]{0.45\linewidth}
        \centering
        \vspace{0pt}
        \includegraphics[width=\linewidth]{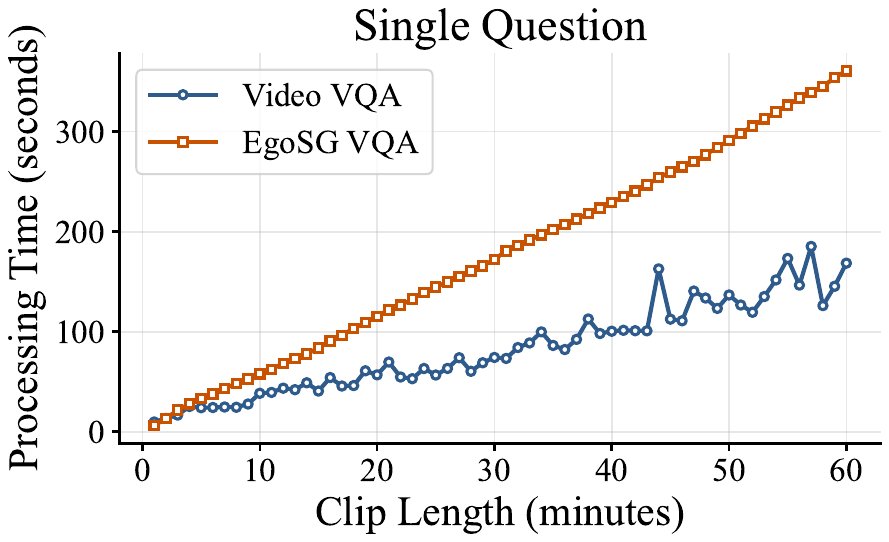}\\[4pt]
        \includegraphics[width=\linewidth]{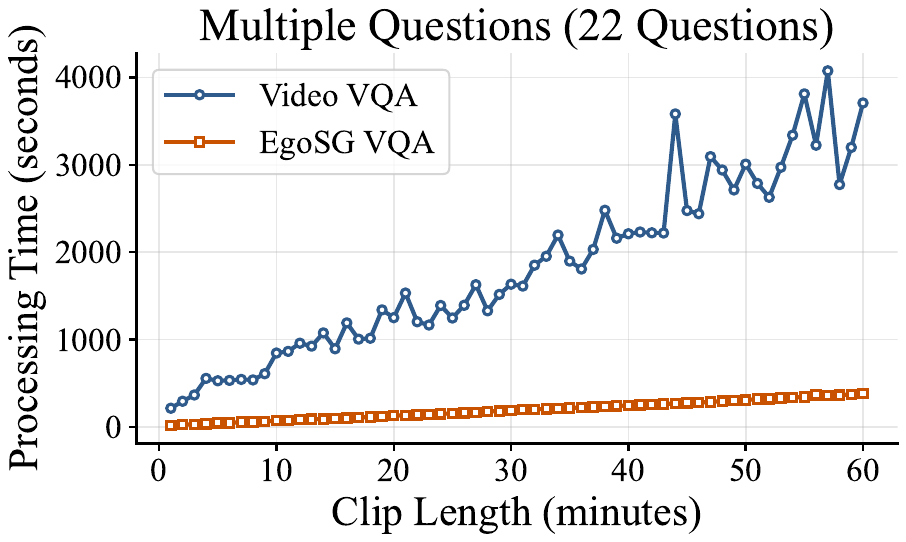}
        \caption{\textbf{Runtime comparison} between raw video and EgoSG.}
        \label{fig:runtime}
    \end{minipage}
    \hfill
    \begin{minipage}[t]{0.45\linewidth}
        \centering
        \vspace{0pt}
        \includegraphics[width=\linewidth]{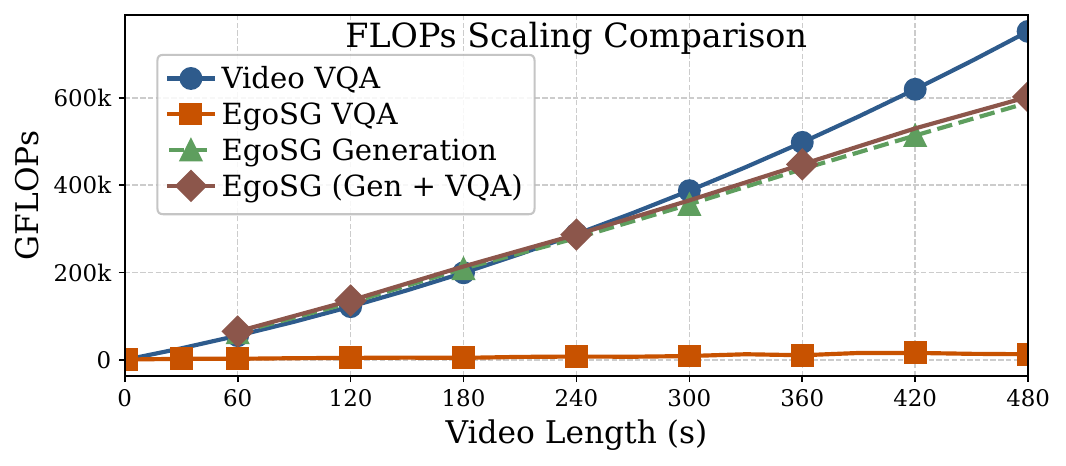}
        \caption{\textbf{FLOPs comparison} between raw video and EgoSG.}
        \label{fig:flops}
        \vspace{10pt}
        \centering
        \includegraphics[width=0.95\linewidth]{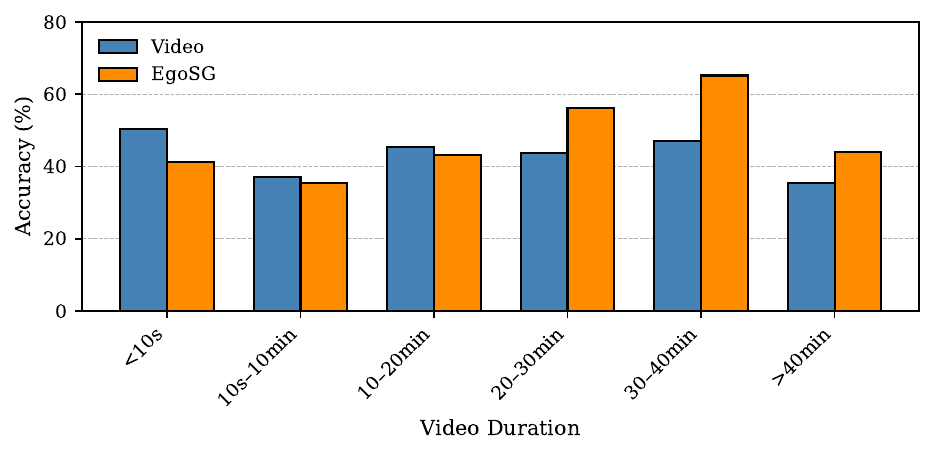}
        \caption{\textbf{Accuracy \textit{vs} video length.} Relationship between model
    accuracy (\%) and video length. }
        \label{fig:length}
    \end{minipage}
\end{figure}

\subsection{Runtime Analysis} 
We report a runtime analysis on Gemini in \cref{fig:runtime}. Importantly, the \textit{EgoSG VQA} curve measures the \emph{end-to-end} cost of our pipeline, i.e., \emph{scene-graph generation + downstream VQA}. EgoSG incurs an upfront graph-generation cost ($\approx$5.7\,s per 1-minute clip) that scales linearly with video length. For a \emph{single} question, this can make EgoSG slower than video-only inference (\cref{fig:runtime}, top). However, graph generation is performed only once per video: the resulting scene graph is a reusable representation that can be queried multiple times without reprocessing the raw video. In the realistic multi-question setting of HD-EPIC (22 questions per video on average), the generation cost is amortized, and EgoSG achieves substantially lower total runtime than video-based VQA (\cref{fig:runtime}, bottom).

The efficiency gains of EgoSG are also reflected in \emph{inference compute}. Video-based MLLMs attend over $T$ visual tokens, with $\mathcal{O}(T^2)$ self-attention cost, whereas EgoSG converts the video into a compact textual representation of $S$ tokens, yielding $\mathcal{O}(S^2)$ complexity with $S \ll T$. This leads to a reduced attention cost, consistent with token trends in \cref{fig:tokens_per_model}. Since API models do not expose FLOPs, we estimate compute using Qwen2.5-VL-3B as an open-source proxy and report GFLOPs in \cref{fig:flops}.
The analysis is performed on a representative video example using a fixed question prompt while varying video duration from 0 to 480 seconds. The figure compares \textit{Video VQA}, \textit{EgoSG Generation}, \textit{EgoSG VQA}, and the \textit{combined} EgoSG pipeline cost (\textit{Gen+VQA}). As video length increases, \textit{Video VQA} scales steeply, while \textit{EgoSG VQA} remains nearly flat, and graph generation grows more slowly. Consequently, the end-to-end EgoSG cost remains lower for long videos, confirming its favorable scaling.

\subsection{Ablations}
\subsubsection{Accuracy \textit{vs} video length.}  \cref{fig:length} illustrates how the performance varies with video length. We observe that EgoSG outperforms the video-based approach on longer videos, where processing the full video becomes challenging for the models. This highlights the robustness of EgoSG in handling extended video sequences compared to conventional frame-based representations.

\begin{figure}[t]
    \centering
    \begin{minipage}[t]{0.49\linewidth}
        \centering
        \includegraphics[width=\linewidth]{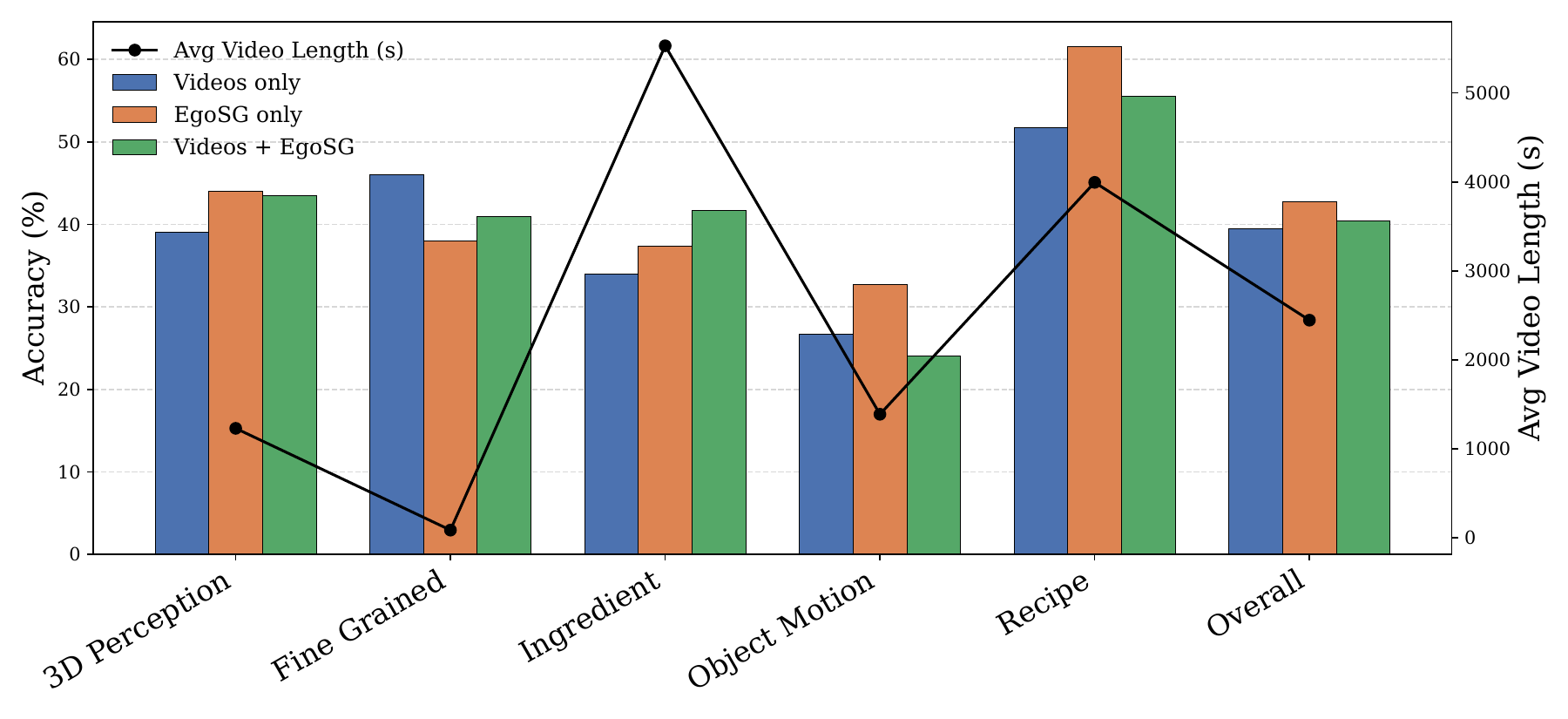}
        \caption{\textbf{EgoSG (w/ and w/o video).} Accuracy (\%) of EgoSG w/ and w/o using the raw video input.}
        \label{fig:video+egosg}
    \end{minipage}
    \hfill
    \begin{minipage}[t]{0.49\linewidth}
        \centering
        \includegraphics[width=\linewidth]{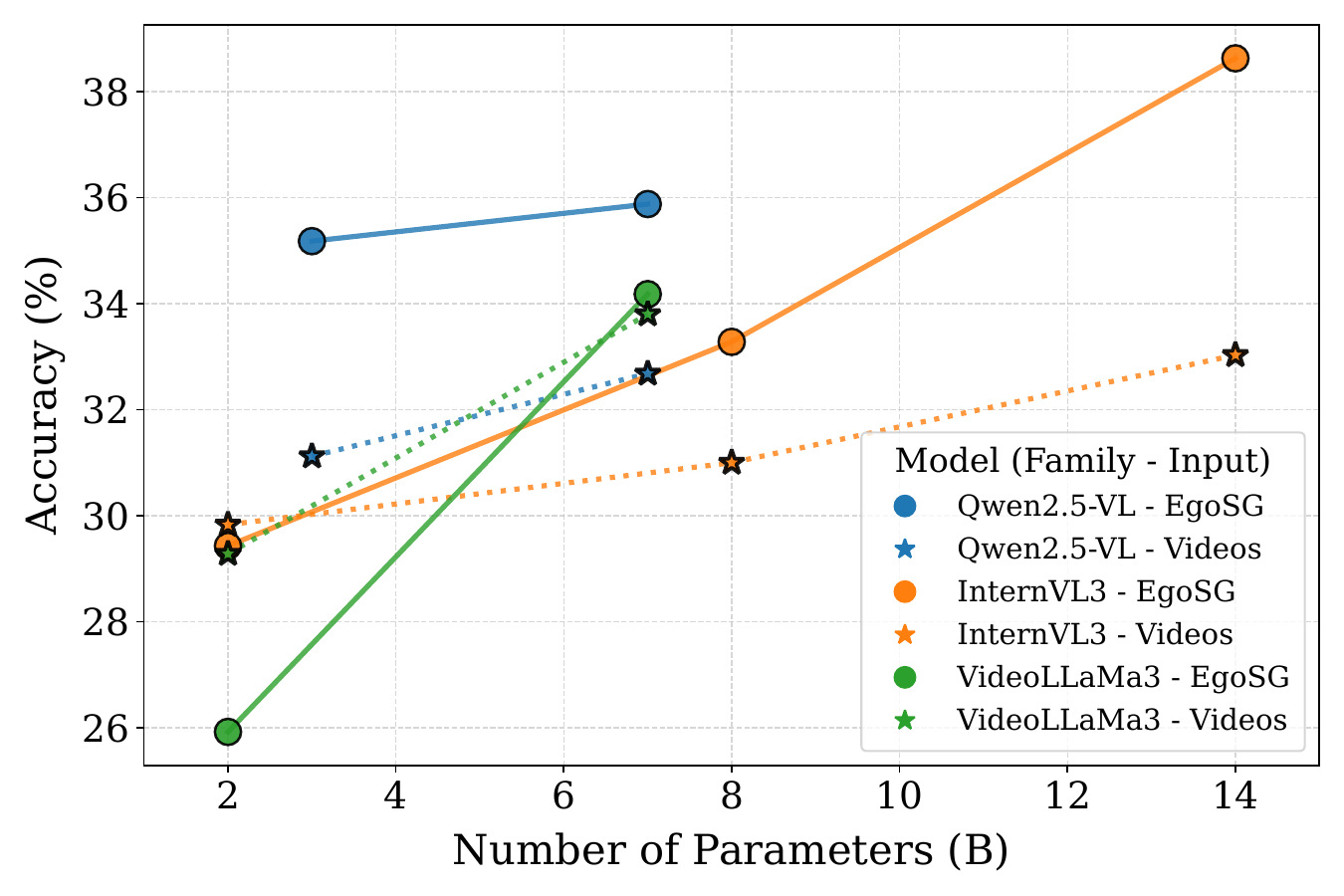}
        \caption{\textbf{Accuracy \textit{vs.} Model Size.} Relationship between model accuracy (\%) and number of parameters in different MLLMs.}
        \label{fig:params}
    \end{minipage}
\end{figure}

\subsubsection{EgoSG (w/ and w/o video).}
\cref{fig:video+egosg} reports per-category results comparing EgoSG-only inference to EgoSG augmented with raw video. The impact of modality fusion varies across categories. In some cases, adding video improves performance (e.g., \textit{Fine Grained} and \textit{Ingredient}), suggesting complementary information between structured scene knowledge and raw visual cues. However, in other categories such as \textit{Object Motion} and \textit{Recipe}, performance decreases, indicating that additional visual tokens may introduce noise, compete for attention, and consume limited context budget, which can ultimately reduce accuracy \cite{deng2025words}.

\begin{figure}[t]
    \centering
    \includegraphics[width=\linewidth]{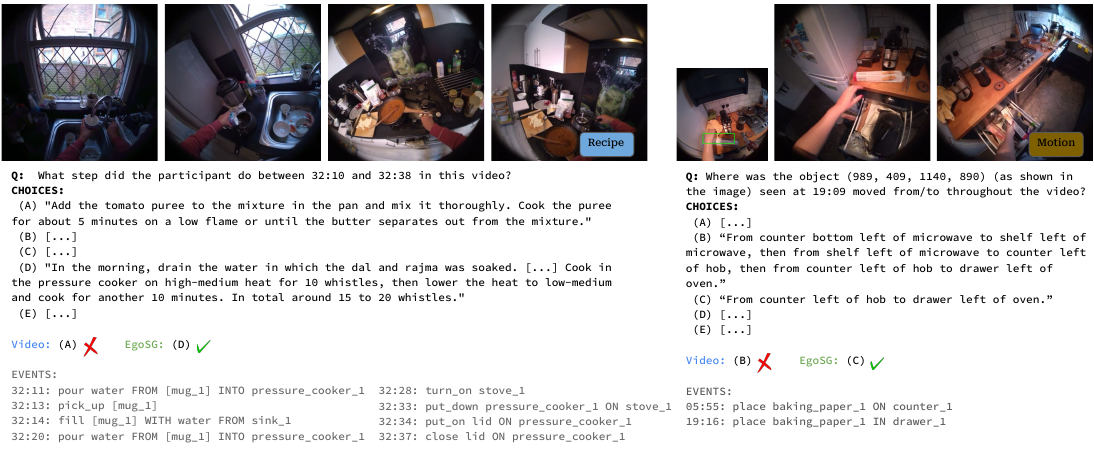}
    \caption{\textbf{Successful Cases.} Instances where EgoSG succeeds and raw video fails.}
    \label{fig:qualitative}
\end{figure}

\begin{figure}[t]
    \centering
    \includegraphics[width=\linewidth]{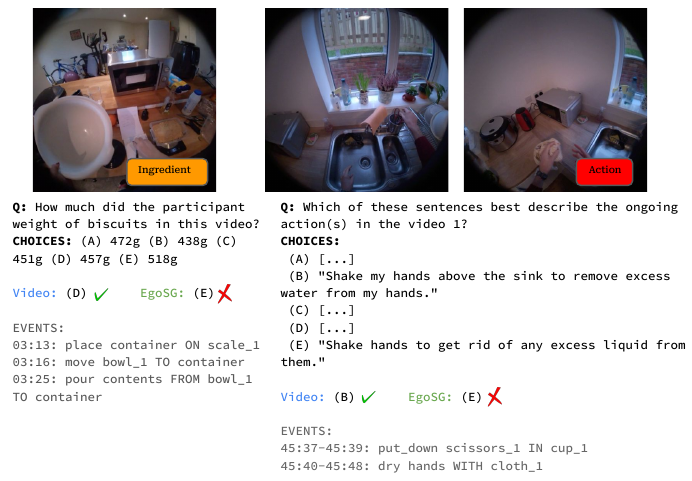}
    \caption{\textbf{Failure cases.} Instances where EgoSG fails and raw video succeeds.}
    \label{fig:failure}
\end{figure}

\subsubsection{Accuracy \textit{vs} number of parameters. }
\cref{fig:params} illustrates how model accuracy varies with the number of parameters.
The relative gain from using EgoSG is more pronounced for larger models.
This trend suggests that small models may lack the capacity to effectively reason over structured representations like scene graphs, likely due to limited exposure to such data during training.

\begin{table}[t]
    \centering  
    \scriptsize
    \caption{\textbf{Ablation on graph structure.} Accuracy (\%) across input formats and task categories. Narrations use ground-truth expert annotations (oracle upper bound). Video and Captions are baselines; Captions are 1-minute free-form descriptions generated directly from video without graph construction. EgoSG is our structured scene-graph event list, and EgoSG$^\dagger$ replaces structured events with 1-minute summaries conditioned on (grounded in) the scene-graph extraction outputs.}
    \label{tab:results_by_format_and_category}
    \begin{tabular}{lccccc|c}
    \toprule
    \textbf{Format} & \textbf{Recipe} & \textbf{Ingredient} & \textbf{Action} & \textbf{3D} & \textbf{Motion} & \textbf{Avg} \\
    \midrule
    \rowcolor{gray!20}
    Narrations  & 80.00 & 54.67 & 85.50 & 45.50 & 32.00 & 59.53 \\
    \midrule
    Video       & 51.75 & 34.00 & \textbf{46.00} & 39.00 & 26.67 & 39.48 \\
    Captions    & \textbf{65.00} & 34.00 & 28.50 & \textbf{40.00} & 30.00 & 39.50 \\
    \midrule
    EgoSG$^\dagger$ & 42.50 & 30.00 & 41.33 & 31.33 & \textbf{63.25} & 41.15 \\
    EgoSG & 44.00 & \textbf{38.00} & 37.33 & 32.67 & 61.50 & \textbf{42.70} \\
    \bottomrule 
    \end{tabular}
\end{table}

\subsubsection{Ablation on graph structure.}
~\cref{tab:results_by_format_and_category} ablates the impact of feeding EgoSG’s graph-based representation (\cref{subsec:schema}) to an MLLM (Gemini Flash 2.0), compared to an unstructured language baseline. As an oracle upper bound, we report \textit{Narrations}, i.e., the dataset’s annotations, which provide dense textual descriptions at an average rate of $\sim$24 sentences per minute (e.g., “\textit{the person picks up a plate from a pile of plates using the right hand}”). We compare against two baselines: \textit{Video}-only, and \textit{Captions}, where the same MLLM generates independent 1-minute descriptions directly from the video (i.e., it is prompted to produce text rather than EgoSG graphs). Our proposed representation, {EgoSG}, achieves the strongest average performance, outperforming both baselines. We also evaluate {EgoSG$^\dagger$}, a variant that replaces the structured graph-based format with 1-minute natural-language summaries generated by our pipeline alongside the scene graphs (see \cref{subsec:implementation}). Importantly, \textit{EgoSG$^\dagger$} summaries are graph-grounded: they are generated in the same prompt as scene-graph extraction, and thus are produced given the predicted entities, relationships and timestamped events, unlike \textit{Captions}, which are generated directly from video without constructing any graph. Overall, {EgoSG} performs best, while {EgoSG$^\dagger$} remains competitive and consistently surpasses both baselines, making it a practical alternative under token-budget constraints.

\subsection{Qualitative Results}
\cref{fig:failure,fig:qualitative} provide qualitative examples of EgoSG, illustrating both its strengths and limitations.
\cref{fig:qualitative} shows cases where EgoSG surpasses the video-only baseline: in a long \textit{Recipe} video (over 3,000s), its compact graph representation captures essential procedural steps that frame-based methods fail to track, and in the \textit{Object Motion} category, it correctly infers object relocation using spatial relations. \cref{fig:failure} shows two failure cases consistent with our quantitative analysis: in an \textit{Ingredient} example, the graph omits visual details such as numbers on a scale, and in an \textit{Action} example it encodes the general “drying hands” motion but misses the precise positioning above the sink, reflecting the limited visual granularity of symbolic abstractions.

\section{Conclusion}
We introduced EgoSG, a method for egocentric VQA that leverages scene graphs for improving MLLMs' reasoning.
EgoSG constructs structured representations of both the environment and the user's interactions, enabling MLLMs to perform reasoning on a more structured input than raw video frames.
Compared to conventional video processing, EgoSG is highly token-efficient, making it particularly well-suited for analyzing long-form videos.
Our proposed framework relies on scene graphs with open-vocabulary capabilities, which currently require a powerful model (Gemini Flash 2.0) to generate reliably.
This dependency constitutes a current limitation of our approach. As open-source MLLMs continue to advance, more lightweight alternatives for constructing egocentric scene graphs will be key to enabling scalable and efficient VQA.

\bibliographystyle{splncs04}
\bibliography{main}

\clearpage
\appendix
\section{Additional Analysis}
\label{sec:add_analysis}

\subsection{EgoSG Generation Quality}
\label{subsec:supp_stats}
We analyze the quality of the generated EgoSGs to highlight both their descriptive strengths and current limitations.

\subsubsection{Preposition Grounding Analysis.}
\cref{tab:preposition_stats_supp} summarizes how well binary relationships ($E_B$), typically expressed via prepositions, are grounded to known elements in the node set 
$N$. For each relationship, we evaluate whether the entities returned by the MLLM are present in the predefined set of nodes $N$. Prepositions like \texttt{ON} and \texttt{FROM} show strong grounding performance, with most arguments successfully resolved to existing node IDs. In contrast, prepositions such as \texttt{WITH} and \texttt{UNDER} show high failure rates (43.5\% and 74.5\% respectively), often due to the MLLM’s inability to identify tools or instruments as discrete, trackable objects in the scene.

\begin{table}[t]
\centering
\caption{\textbf{Breakdown of common action prepositions in the generated EgoSGs.} The table shows total counts, successful relation to environmental (\texttt{env}) and dynamic (\texttt{dyn}) objects, and the corresponding failure rate.}
\label{tab:preposition_stats_supp}
\sisetup{group-separator={,}, group-minimum-digits=4}
\begin{tabular}{l S[table-format=5.0] S[table-format=4.0] S[table-format=4.0] S[table-format=2.1]}
\toprule
\textbf{Preposition} & {\textbf{Total}} & {\textbf{Env}} & {\textbf{Dyn}} & {\textbf{Fail Rate (\%)}} \\
\midrule
ON                   & 10834 & 6900 & 2968 & 8.9 \\
FROM                 & 10825 & 6404 & 3014 & 13.0 \\
WITH                 & 5427  & 52   & 3015 & 43.5 \\
IN                   & 4657  & 2187 & 1804 & 14.3 \\
INTO                 & 1416  & 513  & 729  & 12.3 \\
TO                   & 955   & 367  & 400  & 19.7 \\
OF                   & 335   & 99   & 234  & 0.6 \\
UNDER                & 161   & 39   & 2    & 74.5 \\
AND                  & 113   & 0    & 110  & 2.7 \\
\bottomrule
\end{tabular}
\end{table}
\paragraph{Graph Quality Analysis. } 

\begin{table}[t]
\centering
\caption{\textbf{Graph quality analysis} on 5 representative video clips.}
\label{tab:manual_error_analysis}
\begin{tabular*}{0.8\textwidth}{@{\extracolsep{\fill}} l r @{}}
\toprule
\textbf{Error Category} & \textbf{Accuracy (\%) (Correct/Total)} \\
\midrule
\rowcolor[HTML]{F5F5F5}
\multicolumn{2}{l}{\textbf{Nodes ($N$) Errors}} \\
\addlinespace[0.6ex]
Missing Nodes & 16 missing \\
Wrong Node Attributes & 5.0\% (5/101) \\
\midrule
\rowcolor[HTML]{F5F5F5}
\multicolumn{2}{l}{\textbf{Binary Relationships ($E_B$) Errors}} \\
\addlinespace[0.6ex]
Wrong Spatial Relations & 5.0\% (5/101) \\
\midrule
\rowcolor[HTML]{F5F5F5}
\multicolumn{2}{l}{\textbf{Action Hyperedges ($A$) Errors}} \\
\addlinespace[0.6ex]
False Action Hyperedges & 15.5\% (50/322) \\
Malformed Actions & 9.9\% (32/322) \\
Actions with Wrong Nodes & 8.1\% (26/322) \\
Missing Actions & 13 missing \\
Timestamp Errors & 3.1\% (10/322) \\
\bottomrule
\end{tabular*}
\end{table}

EgoSG scene graphs capture \texttt{ENVIRONMENTAL ELEMENTS}, \texttt{DYNAMIC OBJECTS}, and user-object interactions in the form of \texttt{EVENTS} (see Section 3.3 and Fig. 3 in the main paper). To evaluate the quality of our generated EgoSGs, we conducted a manual inspection on 5 random video clips ranging from 3.0 to 3.8 minutes each (total: 17.2 minutes), containing in total 322 Action Hyperedges $\mathcal{A}$ and 101 \textit{nodes} $\mathcal{N}$ (53 \texttt{Environment Elements} and 48 \texttt{Dynamic Objects}). \cref{tab:manual_error_analysis} summarizes the key error patterns observed.
\textbf{For Nodes}, the system consistently misses interacted objects that are in the video but are not listed in the graph (16 missing in total across clips), such as tools or utensils used in actions but not cataloged. \textit{Wrong Node Attributes} (5.0\%) reflect inaccurate \textit{visual properties} descriptions (colors, materials, labels). \textbf{For Binary Relationships} describing interactions between nodes, \textit{Wrong Spatial Relations} (5.0\%) include errors in the description of relations between nodes and general spatial understanding errors, such as misidentifying location-based relationships or incorrectly describing object positioning within the scene.
\textbf{For Action Hyperedges}, \textit{False Action Hyperedges} represent the most significant challenge (15.5\%), where the MLLM generates plausible but incorrect interaction descriptions (e.g., recording ``pick\_up bottle" when the person actually put it down). \textit{Malformed Actions} (9.9\%) involve missing object references, such as ``move TO" without specifying the destination. \textit{Actions with Wrong Nodes} (8.1\%) occur when the action type is correct but references incorrect entities (e.g., ``wash plate\_1" instead of ``wash bowl\_2"). \textit{Missing Actions} represent significant interactions that occurred but were not captured in the scene graph (13 missing in total). \textit{Timestamp Errors} (3.1\%) indicate events recorded with a timing offset of more than 2 seconds from when they actually occurred. Overall, on average, $5\%$ of the nodes or relationships contain errors, and the 
per-category error rate for actions is $9\%$. In addition, $16/117$ objects ($13.7\%$) 
are missing and $13/335$ actions ($3.9\%$) are undetected.
 
These statistics are obtained from a rigorous manual inspection, a process that is complex and time-consuming due to the density of information in long-form egocentric videos. Precisely measuring how generation errors affect VQA outcomes is challenging, as the impact of a single graph inaccuracy depends on several factors, such as the semantic grounding of the question within the graph-based reasoning process, and whether the question actually relies on the portion of the scene affected by the error. Moreover, manually tracing these dependencies across long, visually dense egocentric videos is infeasible in practice.
Critically, this work does not focus on optimizing scene graph generation quality, but on demonstrating that structured intermediate abstractions encoding entities, relations, and events are effective representations for downstream reasoning. The strong results achieved despite generation imperfections underscore the robustness and utility of the EgoSG framework.

\subsection{Input Processing Strategies}
\label{sec:supp_context}
\begin{table}[t]
\centering
\caption{\textbf{Frame sampling limits.} Maximum number of frames fed to each model per question.}
\label{tab:video_adaptation}
\sisetup{group-separator={,}, group-minimum-digits=4}
\small
\begin{tabular}{l S[table-format=7.0]}
\toprule
\textbf{Model} & \textbf{Max Input Frames} \\
\midrule
Gemini Flash 2.0 &  3900 \\
Qwen2.5-VL       &  512 \\
VideoLLaMA3      &  180 \\
InternVL3        &  32 \\
\bottomrule
\end{tabular}
\end{table}

\begin{table}[t]
\centering
\caption{\textbf{Compression strategies for EgoSG inputs.} The ``Summaries Used'' column indicates how often the full graph was replaced by its summary version.
$^{\dagger}$: For InternVL3, 150 questions exceed the strict context limit even after summarization and cannot be processed; these instances are treated as incorrect predictions in the accuracy reported in the main paper.}
\label{tab:sg_adaptation}
\sisetup{group-separator={,}, group-minimum-digits=4}
\small
\begin{tabular}{l S[table-format=7.0] l l}
\toprule
\textbf{Model} & {\textbf{Context Window}} & \textbf{Summaries Used}  \\
\midrule
Gemini Flash 2.0 & 1048576 & 0 / 1,250 (0.0\%) \\
Qwen2.5-VL       & 32768   & 154 / 1,250 (12.3\%)  \\
VideoLLaMA3      & 32768   & 154 / 1,250 (12.3\%)  \\
InternVL3        & 8192    & 565 / 1,100$^{\dagger}$ (51.4\%) \\
\bottomrule
\end{tabular}
\end{table}

To accommodate the varying capacities of different MLLMs, particularly for questions involving long or multi-video inputs, we employ model-specific frame sampling strategies. We initially sample all videos at a uniform rate of 1 FPS, applying further uniform subsampling whenever the resulting frame count exceeds the maximum input limits per question reported in \cref{tab:video_adaptation}. For Gemini Flash 2.0, we utilize its high input capacity to process up to 3,900 frames, effectively covering over an hour of footage without further reduction.
In contrast, open-source models require stricter limits:  VideoLLaMA3 and InternVL3 are restricted to 180 and 32 frames, respectively, according to the capacity of their context length. For Qwen2.5-VL, we reduce the input from the standard limit of 768 to 512 frames to accommodate GPU memory limitations.
Consequently, for long-form videos exceeding these budgets, baseline models are forced to severely subsample frames.
In contrast, our approach encodes the entire video duration, enabling the downstream model to reason over the full temporal context equivalent to 1 FPS resolution.

For EgoSG inputs, we primarily employ the complete structured graph. However, for models with limited context windows, this dense representation can occasionally exceed the token budget when processing long-form videos. In such instances, we employ a fallback strategy: replacing the structured graph with the concatenated natural language summaries generated for each video clip (see the ``Summary'' field in the prompt definitions in \cref{sec:prompts}). This provides a coarser narrative of the video that fits within stricter token limits. \cref{tab:sg_adaptation} details the frequency with which this adaptation was applied.
Importantly, as demonstrated in the ablation study in the main paper (Table 4), substituting the full graph with these summaries results in only a minor degradation in the overall accuracy.
\begin{figure}[t]
    \centering
    \includegraphics[width=\linewidth]{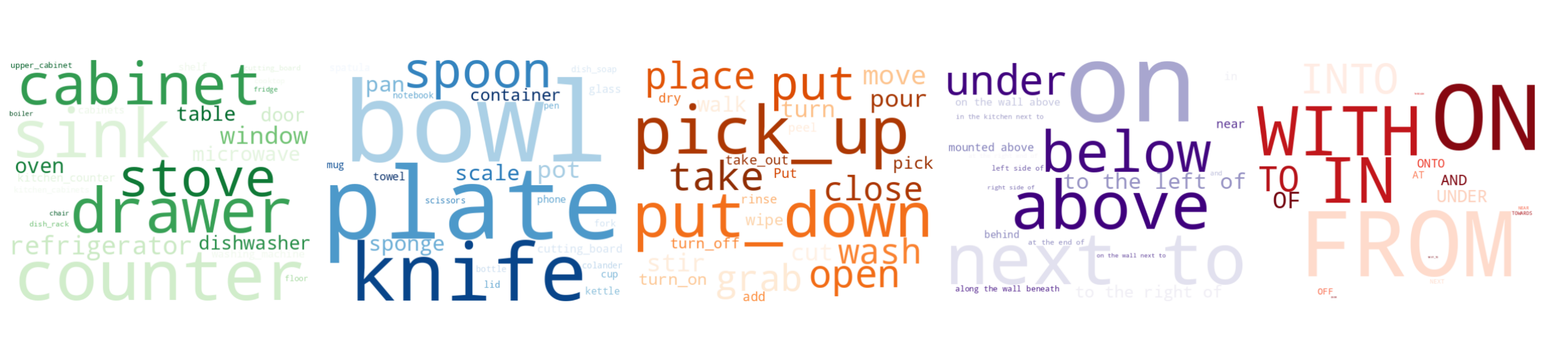}
    \caption{\textbf{EgoSG components distribution.} From left to right: wordles of (1) environment elements, (2) dynamic objects, (3) action types, (4) location-based relations and (5) action prepositions.}
    \label{fig:words}
\end{figure}

\subsection{EgoSG Components}
We present in \cref{fig:words} the top 25 most frequent items for the following categories: (1) environment elements, (2) dynamic objects, (3) action types, (4) location-based relations, and (5) action prepositions.

\section{Results on the Complete Evaluation Subset}
\label{sec:full_subset_results}

In the main paper, we focus our evaluation on the five HD-EPIC VQA categories that most directly require long-term temporal and spatio-temporal reasoning, excluding \textit{Nutrition} and \textit{Gaze}. For completeness, \cref{tab:complete_subset_results} reports results on the complete evaluation subset, including all seven selected categories. This subset contains 1,500 questions in total.

\begin{table}[t]
\centering
\caption{\textbf{Results on the complete evaluation subset.} Accuracy (\%) comparison between EgoSG and raw video across all selected categories, including \textit{Nutrition} and \textit{Gaze}. The \textit{Overall} column reports the mean accuracy across categories, following the reporting protocol used in the main paper. Best results are \textbf{bolded}.}
\label{tab:complete_subset_results}
\resizebox{\textwidth}{!}{
\setlength{\tabcolsep}{4pt}
\begin{tabular}{lccccccc>{\columncolor[gray]{0.95}}c}
\toprule
\textbf{Input} &
\textbf{Recipe} &
\textbf{Ingred.} &
\textbf{Nutrition} &
\textbf{Action} &
\textbf{3D Perc.} &
\textbf{Object Mot.} &
\textbf{Gaze} &
\textbf{Overall} \\
\midrule
Video & 51.75 & 34.00 & \textbf{32.67} & \textbf{46.00} & 39.00 & 26.67 & \textbf{34.00} & 37.73 \\
EgoSG & \textbf{61.50} & \textbf{37.33} & 24.67 & 38.00 & \textbf{44.00} & \textbf{32.67} & 29.00 & \textbf{38.17} \\
\bottomrule
\end{tabular}
}
\end{table}

\section{Prompts for EgoSG Generation}
\label{sec:prompts}
To generate structured scene graphs from egocentric video clips, we adopt a sequential prompting strategy designed to ensure temporal consistency and reduce redundancy across clips.
Each prompt includes a description of the desired output format (see \cref{fig:scene_graph_format}), organized into four main sections:
\begin{enumerate}
    \item \textbf{Environment Elements:} Unique identifiers, spatial relationships, and visual descriptors.
    \item \textbf{Dynamic Objects:} Positioning and appearance details for movable items.
    \item \textbf{Events:} Timestamped actions, formatted using standardized capitalized prepositions (e.g., \texttt{PLACE ON}, \texttt{PICK UP FROM}).
    \item \textbf{Summary:} A textual description summarizing the main activities in the clip.
\end{enumerate}
This structured format enables consistent parsing and seamless integration into our downstream video reasoning pipeline.

\subsubsection{First Clip Prompt.}
For the first clip in each sequence, we use a comprehensive prompt (see \cref{fig:first_clip_prompt}) to initialize the scene graph. 
The model is instructed to extract all graph elements from scratch, and we provide an example scene graph (\cref{fig:scene_graph_example}) as a reference.

\subsubsection{Subsequent Clips Prompt.}
For each following clip, the prompt includes contextual information extracted from previous clips (see \cref{fig:other_clip_prompt}).
Specifically, it provides the full set of previously identified environment elements and dynamic objects.
To manage token limits while preserving context, we only include the list of events from the immediately preceding clip, while earlier clips are summarized using their generated high-level summaries (for the second clip, only the events from the first are included).
This design ensures the model retains relevant temporal and spatial context while staying within token limits.
The model is then asked to identify new or changed entities and avoid repeating previously detected elements.
This strategy helps maintain coherence across time while keeping the prompts concise and efficient.

\begin{figure}[t]
\begin{tcolorbox}[title=Prompt Template for Initial Clip Processing, colback=gray!10, colframe=black, sharp corners, boxrule=0.5pt]
\ttfamily
You are processing the first clip of an egocentric activity video.\\
Please carefully observe the video clip and extract the following:\\
1) Environment Elements: Large and medium appliances, built-in storage structures (e.g., cabinets, drawers), work surfaces (e.g., counters, tables), and fixtures (e.g., sinks, wall-mounted switches). List up to 10 items, describing their location relative to other environment elements, and key visual traits.\\
2) Dynamic Objects: Movable objects that the user interacts with during this clip and can be manipulated to change their location within the environment. List up to 10 items, describing their initial position relative to environment elements or other known objects, and appearance or label.\\
3) Events and Interactions: Key events and interactions the user performs in this clip, each with a timestamp. List up to 15 relevant events.\\
4) Brief summary describing the main actions of the clip.\\
---\\
\textbf{[SCENE GRAPH OUTPUT FORMAT (see \cref{fig:scene_graph_format})]}\newline
---\\
Example:\\
\textbf{[EXAMPLE SCENE GRAPH (see \cref{fig:scene_graph_example})]}\newline
---\\
Think step by step, then output only the three lists and the summary as described.\\
\end{tcolorbox}
\caption{\textbf{Prompt template for processing the first clip in an egocentric video sequence. }The template establishes the foundational scene graph structure by identifying environment elements, dynamic objects, and temporal events.}
\label{fig:first_clip_prompt}
\end{figure}

\begin{figure}[t]
\begin{tcolorbox}[title=Prompt Template for Subsequent Clip Processing, colback=gray!10, colframe=black, sharp corners, boxrule=0.5pt]
\ttfamily
You are processing a new clip from an egocentric activity video.\\
Below are the accumulated Environment Elements and Dynamic Objects identified from all previous clips:\\
\textbf{[ACCUMULATED ENVIRONMENT ELEMENTS]}\\
\textbf{[ACCUMULATED DYNAMIC OBJECTS]}\\
Below are the Summaries of older clips and the Events from the most recent previous clip:\\
\textbf{[SUMMARIES FROM OLDER CLIPS]}\\
\textbf{[EVENTS FROM MOST RECENT CLIP]}\\
---\\
Your task is to analyze the new clip and extract the following:\\
1) NEW Environment Elements that were NOT previously listed. These are large and medium appliances, built-in storage structures (e.g., cabinets, drawers), work surfaces (e.g., counters, tables), and fixtures (e.g., sinks, wall-mounted switches). List up to 10 items, describing their location relative to other stationary objects, and key visual traits. Leave this section empty if there are no additional items.\\
2) NEW Dynamic Objects that were NOT previously listed. These are movable objects that the user interacts with during this clip and can be manipulated to change their location within the environment. List up to 10 items, describing their initial position relative to environment elements or other known objects, and appearance or label. Leave this section empty if there are no additional items.\\
3) Events and Interactions: Key events and interactions the user performs in this clip, each with a timestamp. List up to 15 relevant events. You may reference IDs from previous Environment Elements and Dynamic Objects where appropriate.\\
4) Brief summary describing the main actions of the current clip.\\
---\\
\textbf{[SCENE GRAPH OUTPUT FORMAT (see \cref{fig:scene_graph_format})]}\newline
---\\
Think step by step, then output only the three lists and the summary as described.
\end{tcolorbox}
\caption{\textbf{Prompt template for processing clips from the second on}, using accumulated context from all previous clips while managing prompt length through selective inclusion of recent events and older summaries.}
\label{fig:other_clip_prompt}
\end{figure}

\begin{figure}[t]
\begin{tcolorbox}[title=Structured Output Format Specification, colback=gray!10, colframe=black, sharp corners, boxrule=0.5pt]
\ttfamily
Format your response as shown below:\\
ENVIRONMENT ELEMENTS:\newline
[element\_id]: [location description] - [visual description]\\
…\\
DYNAMIC OBJECTS:\newline
[object\_id]: [initial position description] - [appearance description or label]\\
…\\
EVENTS:\newline
[timestamp]s: [verb] [object] [PREPOSITION secondary\_object]*\\
...\\
SUMMARY:\\
...\\
---\\
For ENVIRONMENT ELEMENTS and DYNAMIC OBJECTS identification, assign unique and consistent IDs using underscore notation.\\
In the EVENTS:\\
- When appropriate, include one or more capitalized prepositional phrases (e.g., ON, IN, FROM, TO, WITH) to indicate spatial or instrumental relationships with other DYNAMIC OBJECTS or ENVIRONMENT ELEMENTS involved in the action.\\
- It is acceptable to refer to subcomponents (e.g., button ON boiler) or contents (e.g., milk FROM carton) even if these are not separately listed in the DYNAMIC OBJECTS section, as long as they represent an identifiable part of the interaction.\\
\end{tcolorbox}
\caption{\textbf{Structured output format specification} that ensures consistent formatting across all clip processing stages.}
\label{fig:scene_graph_format}
\end{figure}

\begin{figure}[t]
\begin{tcolorbox}[title=Example Scene Graph Output for Tea Preparation Scenario, colback=gray!10, colframe=black, sharp corners, boxrule=0.5pt]
\ttfamily
ENVIRONMENT ELEMENTS\\
refrigerator\_1: along the left wall -- tall white appliance with a vertical handle\\
main\_counter: along the wall beneath upper\_cabinet\_1 -- light wooden surface with a white tiled backsplash\\
upper\_cabinet\_1: mounted above main\_counter -- cream-colored wood with glass panel doors\\
drawer\_1: below main\_counter -- cream-colored front with silver handle\\
kettle\_1: on main\_counter near sink\_1 -- silver electric kettle with black base\\
sink\_1: at the right end of main\_counter -- stainless steel with a tall faucet\\\\
DYNAMIC OBJECTS\\
mug\_1: inside upper\_cabinet\_1 -- white ceramic with floral print\\
tea\_bag\_1: inside drawer\_1 -- small paper envelope labeled "Green Tea"\\
water\_bottle\_1: inside refrigerator\_1 -- transparent plastic with blue cap\\\\
EVENTS\\
3s: open upper\_cabinet\_1\\
4s: pick\_up mug\_1 FROM upper\_cabinet\_1\\
5s: put\_down mug\_1 ON main\_counter\\
7s: open drawer\_1\\
8s: pick\_up tea\_bag\_1 FROM drawer\_1\\
10s: put\_down tea\_bag\_1 INTO mug\_1\\
12s: open refrigerator\_1\\
13s: grab water\_bottle\_1 FROM refrigerator\_1\\
15s: close refrigerator\_1\\
17s: pour water FROM water\_bottle\_1 INTO kettle\_1\\
20s: turn\_on kettle\_1\\\\
SUMMARY:\\
The user prepares to make tea by placing a mug on the counter, adding a tea bag, and filling a kettle with water from the refrigerator. They then turn the kettle on to heat the water.
\end{tcolorbox}
\caption{\textbf{Example scene graph output included in the first clip prompt template}, provided to the MLLM to indicate the structured format illustrating proper identification of environment elements, dynamic objects, and temporal events with spatial relationships.}
\label{fig:scene_graph_example}
\end{figure}

\section{Ablation on Clip Duration}
\label{sec:ablation-length}
We study the effect of input clip duration when generating EgoSGs. Specifically, we evaluate four clip lengths: 30 seconds, 1 minute (default), 
2 minutes, and 5 minutes, and report the results in \cref{tab:ablation_duration}.

The results show that 1-minute clips achieve the best overall performance. 
Using shorter clips (30 seconds) leads to a larger number of temporal segments per video, increasing update frequency and often fragmenting continuous actions across multiple clips. This fragmentation slightly reduces accuracy.
In contrast, longer clips (2–5 minutes) contain substantially more objects and actions within the same temporal window. The model must therefore summarize a larger amount of visual information, which leads to coarser scene-graph representations and lower accuracy across all evaluation
categories.

\begin{table}[t]
\centering
\caption{\textbf{Ablation on clip duration.} Accuracy (\%) for EgoSG generation with different input clip lengths. Best result per category is \textbf{bolded}.}
\begin{tabular}{l|cccccc}
\toprule
\textbf{Input} &
\textbf{Recipe} &
\textbf{Ingred.} &
\textbf{Action} &
\textbf{3D Perc.} &
\textbf{Object Mot.} &
\textbf{Overall} \\
\midrule
30 sec
& 57.75 & \textbf{38.00} & 36.00 & \textbf{44.00} & 30.67 & 41.28 \\
\textbf{1 min}
& \textbf{61.50} & 37.33 & \textbf{38.00} & \textbf{44.00} & \textbf{32.67} & \textbf{42.70} \\
2 min
& 54.25 & 36.33 & 34.00 & 41.00 & 27.33 & 38.58 \\
5 min
& 41.00 & 28.00 & 28.00 & 36.00 & 25.33 & 31.67 \\
\bottomrule
\end{tabular}
\label{tab:ablation_duration}
\end{table}

\end{document}